C.S. BIANCHINI, F. BORGIA, M. CASTELLI

# L'appropriation et les modifications de SignWriting (SW) par des locuteurs de la Langue des Signes Italienne (LIS)[1]

*Résumé*

Les Langues des Signes (LS) n'ont pas développé un système d'écriture historiquement déterminé ; il existe toutefois des systèmes pour les représenter, dont SignWriting (SW) est un exemple performant. Nous mettons ici en évidence la manière dont les locuteurs de la Langue des Signes Italienne (LIS), utilisateurs experts de SW, ont tendance à modifier les glyphes conventionnels de SW pour pouvoir améliorer leur capacité à représenter la forme des signes ou pour annoter des phénomènes linguistiques particuliers. Après avoir identifié ces glyphes, nous montrons les caractéristiques qui les rendent acceptables par tous nos experts. Nous analysons les possibilités d'utilisation de ces glyphes dans l'écriture manuelle et dans l'écriture informatique, en concentrant notre attention sur la réalisation de SWift, un logiciel conçu pour permettre une éventuelle modification des glyphes par les utilisateurs.

*Abstract*

Historically, the various sign languages (SL) have not developed an own writing system; nevertheless, some systems exist, among which the SignWriting (SW) is a powerful and flexible one. In this paper, we present the mechanisms adopted by signers of the Italian Sign Language (LIS), expert users of SW, to modify the standard SW glyphs and increase their writing skills and/or represent peculiar linguistic phenomena. We identify these glyphs and show which characteristics make them "acceptable" by the expert community. Eventually, we analyze the potentialities of these glyphs in hand writing and in computer-assisted writing, focusing on SWift, a software designed to allow users to modify glyphs.

**Introduction**

Les Langues des Signe (LS) sont le moyen de communication utilisé par la plupart des sourds en Italie (Russo Cardona & Volterra, 2007), en France (Garcia & Derycke, 2010a, 2010b) et par le monde ; il s'agit de langues visuo-gestuelles qui, comme une grande partie des langues vocales (LV) du monde (Cardona, 1981 ; Ong, 1982), n'ont pas développé un système d'écriture qui lui soit propre. Ceci permet d'assimiler les LS, dont la Langue des Signes Italienne (LIS), aux langues à tradition exclusivement orale (Di Renzo *et al.*, 2006a, 2006b ; Pizzuto *et al.*, 2006). Affronter la problématique de la représentation des langues sans écriture, quelles qu'elles soient (Garcia, 2010 ; Garcia & Derycke, 2010a), exige de répondre

---

à des questions sur ce qui doit être transposé (signifié ou signifiant ou les deux ?) et par quel moyen y parvenir. L'article de Bianchini *et al.* (2011), dont ce travail est en quelque sorte le prolongement, expose les différents systèmes qui ont été mis en œuvre par les chercheurs du monde entier pour trouver une solution au problème de la représentation des LS. Qu'il s'agisse d'« étiquettes verbales »[2] ou de formes de notation spécialisées (cf. par exemple Stokoe, 1960 ; Prillwitz *et al.*, 1989 ; Newkirk, 1989), ces systèmes présentent de nombreuses limites (Antinoro Pizzuto *et al.*, 2008; Cuxac, 2000 ; Cuxac & Antinoro Pizzuto, 2010 ; Fabbretti & Pizzuto, 2000 ; Pizzuto *et al.*, 2006 ; Pizzuto & Pietrandrea, 2001 ; Russo, 2004, 2005), dont :

> « Grande complexité dans l'écriture mais aussi dans la lecture ; manque de connexion avec la forme effective du signe ; grande difficulté, voir impossibilité, de transposer des discours en signes ; omission de parties importantes des signes, telle que les composantes non manuelles ; difficultés de mémorisation des critères utilisés pour transposer les signes ; quasi impossibilité de réutiliser les donnés transcrites dans des buts différents de ceux pour lesquels la transcription a été effectuée ; difficultés à rendre la dynamique des signes et l'utilisation particulière de l'espace de signation. » (Bianchini *et al.*, 2011 : 72-73)

Le système de représentation *SignWriting* (SW), développé par Sutton (1995), semble être capable d'affronter le défi d'une mise « noir sur blanc » (Pennacchi, 2008) des LS. Il s'agit d'un système souvent considéré *featural* (Daniels, 1990 ; plus spécifiquement sur SW, Martin, 2000) ou *alphabétique*[3] - non latin -, qui peut être considéré comme un hybride de la notation spécialisée et du dessin (Bianchini *et al.*, 2011 ; Gianfreda *et al.*, 2009), car il est fondé sur un ensemble de symboles conventionnels appelés *glyphes* qui représentent chaque composante du signe (manuelle et non manuelle, comme le regard ou les mouvements du corps) et qui sont placés dans un espace bidimensionnel qui reproduit, en le réduisant à un plan, l'espace de signation tridimensionnel (cf. Fig.1). Cette organisation lui permet de représenter tout type de signe, isolé ou en contexte, que ce signe constitue une Unité Lexématique (Cuxac & Antinoro Pizzuto, 2010) ou une Structure de Grande Iconicité (Cuxac, 2000), termes que nous allons expliquer dans le prochain paragraphe.

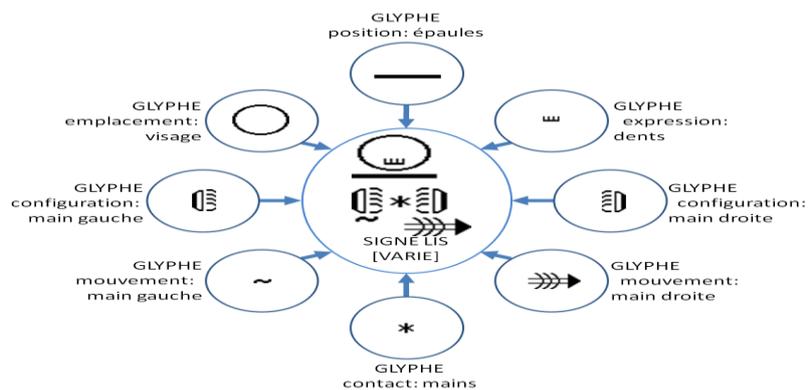

*Fig. 1 - Le SignWriting permet de représenter un signe grâce à l'agencement de glyphes (qui en symbolisent les différents éléments) dans un espace bidimensionnel (qui remplace l'espace de signation tridimensionnel). Dans l'exemple : [VARIE], 'de nombreuses entités différentes', en LIS.*

---

[2] Dans les LV, les gloses servent à exprimer dans une langue de référence le signifié d'un signifiant exprimé dans une autre langue (par une représentation orthographique ou phonétique) ; en LS, les « gloses » sont utilisées pour exprimer le signifiant d'un Signe, mais très souvent le signifiant, donc la forme du Signe, n'est en aucune façon représenté. Il ne s'agit donc pas, comme pour les LV, d'une correspondance forme-signifié, mais de la seule expression du signifié (cit. Bianchini, in press). Nous préférons donc, en accord avec Cuxac & Antinoro Pizzuto, 2010 et Pizzuto *et al.*, 2006, définir ces « gloses » comme simples « étiquettes verbales ».

[3] Nous avons nous même longtemps considéré SW comme alphabétique (cf. Gianfreda *et al.*, 2009 ; Bianchini *et al.*, 2011), pour ensuite le définir comme un système basé sur la représentation de la trace dessinée par le signe dans l'espace, difficilement comparable avec les systèmes de représentation des LV existants (Bianchini, 2016)

# 1. Le SignWriting pour écrire et transcrire des textes en LIS

Le SignWriting a été créé en 1974 par Valérie Sutton, chorégraphe et inventrice d'un système de notation de la danse, en collaboration avec des chercheurs de l'Université de Copenhague [1]. À la fin des années '80, Sutton fonde le D.A.C. (Deaf Action Committee for SignWriting) et le système commence à se diffuser à partir de ce moment-là.

Dans la première moitié des années 2000, SW est adopté par le laboratoire Sign Language and Deaf Studies[4] de l'ISTS-CNR comme système de représentation de la LIS, et devient tant un instrument qu'un sujet de recherche ; leur première approche visait la transcription de signes, afin de tester la capacité de SW à fournir une alternative valable aux systèmes de notation linéaire (étiquettes verbales ou symboles) ainsi qu'aux dessins de signes. La facilité d'apprentissage du système et sa flexibilité ont vite permis aux membres du SLDS de transposer de façon fort détaillée des histoires en signes, exécutés en modalité face-à-face (FàF) et en enregistrements vidéo. Rapidement, l'attention s'est déplacée vers l'utilisation de SW comme système d'écriture de textes conçus et exprimés directement en modalité écrite (Bianchini *et al.*, 2011 ; Di Renzo *et al.*, 2006a, 2006b, 2006c ; Gianfreda *et al.*, 2009), pour permettre la constitution de longs textes exprimés en LIS-Écrite ; il convient ici de souligner la nouveauté de cette approche, puisque les LS sont des langues exclusivement orales et qu'aucun système de notation pris en considération précédemment par le SLDS ne semblait capable de permettre la conception de textes écrits, et non plus seulement transcrits, à partir d'une expression FàF.

L'expérimentation répétée de SW au sein du SLDS, tant comme système de transcription que d'écriture, nous a permis d'obtenir, en 2012, un corpus assez vaste et varié, de textes et de signes isolés. Si l'écriture manuelle est le premier pas de tous nos textes, il y a des cas où les membres du groupe ont décidé de numériser ces données en recourant à un logiciel spécifique, fourni par Sutton, le *SignMaker* [2]. Cette application se base sur l'ensemble des glyphes officiellement reconnus par Sutton, qui sont recueillis dans une liste appelée *International SignWriting Alphabet* (ISWA) ; celui-ci se nommait, en 2004[5], *International Mouvement Writing Alphabet* (IMWA) et il comptait 29 276 glyphes [5].

Il est impossible de traiter ici dans le détail les modifications subies par SW aux cours des années[6] : dans les grandes lignes, il s'agit de différences qui augmentent le nombre de glyphes et donc aussi la précision dans la représentation des composantes des signes ; toutefois, l'organisation structurale même de SW - basée sur des codes d'identification des glyphes qui changeaient radicalement à chaque nouvelle version - rendait difficile la migration informatique des données d'une version à l'autre du logiciel SignMaker. C'est la raison pour laquelle les experts de SW du groupe SLDS préféraient (en 2012, quand notre corpus a été recueilli) continuer à utiliser le SignMaker 1.4 (datant de 2004), en attendant le moment où SW se serait stabilisé, chose qui peut être désormais considérée comme advenue puisque SW a été reconnu, en 2017, par le Consortium Unicode [6]

En 2009, lors de la numérisation des signes écrits manuellement nécessaires à la préparation d'un travail sur les signes présents dans les dictionnaires (Bianchini *et al.*, 2009), nous avons remarqué la présence de nombreux glyphes qu'il était impossible de retrouver dans l'IMWA2004 et donc dans SignMaker, mais dont l'interprétation ne posait aucun problème ni pour nous ni pour les autres utilisateurs de SW avec lesquels nous travaillions. Ceci nous a poussés à approfondir la question, en étendant la recherche de ces glyphes aux autres textes en SW produits par le groupe du SLDS : nous présentons ici les résultats et les conséquences de ce travail. Nous verrons donc en premier lieu quels types de glyphes sont créés par les membres du SLDS et quelles sont leurs caractéristiques communes, pour ensuite trouver des solutions permettant la numérisation de ces glyphes « non conventionnels ».

---

[4] Le Laboratoire Sign Language and Deaf Studies (SLDS), connu jusqu'en 2010 comme Sign Language Laboratory (SLL), a été fondé par Elena Antinoro Pizzuto† et fait partie de l'Institut de Sciences et Technologies de la Cognition (ISTC) du Conseil National de la Recherche (CNR) italien ; en 2012, à l'apogée de la recherche sur SW en Italie, le SLDS comptait, entre personnel fixe et collaborateurs externes, 10 membres dont 6 sourds et 4 entendants.
[5] Nous faisons ici référence à l'IMWA2004, et non aux ISWA2008 [3] et ISWA2010 [4], plus récents, car les membres du SLDS utilisent exclusivement la version 1.4 de SignMaker qui se base sur l'IMWA2004 et qui est installée sur le serveur interne du laboratoire.
[6] Pour une analyse des différences entre les versions de l'ISWA voir Bianchini (2012) et Bianchini et Borgia (2012*,* 2014).

## 2. De SignWriting officiel à SignWriting non conventionnel

L'écriture manuelle, à cause de la grande liberté donnée par la feuille et le papier, permet de créer de nombreux glyphes non conventionnels, comme indiqué par la Fig. 2.

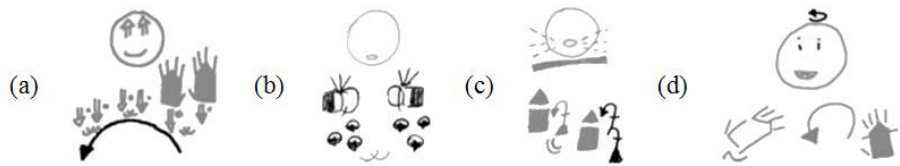

*Fig. 2 – Exemples de la façon dont l'écriture à la main permet d'intégrer aux signes, des glyphes non conventionnels (en noir), inventés par les auteurs des textes.*

Les exemples (a) et (b) mettent en évidence un changement de glyphe très facile à réaliser, c'est-à-dire la possibilité, donnée par l'écriture manuelle, de réduire ou d'agrandir un glyphe représentant le mouvement selon l'ampleur effective de celui-ci. Nous avons donc, en (a), un mouvement semi-circulaire sur le plan horizontal très ample, beaucoup plus que ce que permet de représenter 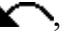, qui est le glyphe équivalent présent dans l'IMWA2004 ; à l'opposé, le glyphe 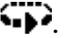 -répété 6 fois dans la partie inférieure de (b) - doit représenter un tout petit mouvement circulaire sur le plan horizontal, qui serait trop ample si l'on utilisait 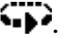.

La situation est plus complexe dans le signe (c), où le but de l'invention du glyphe non conventionnel est celui de répéter deux fois un mouvement de torsion de l'avant-bras, normalement représenté par 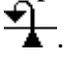. En SW, un glyphe semblable à celui-ci est 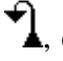, qui indique un mouvement semi-circulaire de la main sur le plan sagittal ; ce mouvement peut être doublé en utilisant le glyphe conventionnel 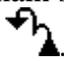. L'auteur prend donc les caractéristiques de 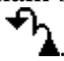 définissant l'utilisation de l'avant-bras (la barre horizontale du glyphe) et les reporte sur le glyphe 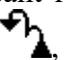, qui représente un double bondissement. Nous obtenons ainsi un glyphe qui, s'il était présent dans l'IMWA2004, aurait été probablement dessiné 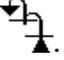.

De même, le glyphe 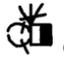 de l'exemple (b) représentant la main avec l'index et le pouce se touchant à peine et les trois autres doigts tendus, ne se retrouve pas dans l'IMWA ; nous y trouvons par contre 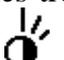, qui peut être utilisé pour représenter cette configuration (nous verrons un peu plus loin pourquoi le glyphe non conventionnel est traversé par des traits horizontaux contrairement au glyphe officiel), mais ce glyphe représente le poing presque fermé (bien qu'arrondi) et les trois doigts tendus, ce qui n'est pas exactement la même chose. L'auteur a donc décidé de redessiner le glyphe pour le rendre plus conforme à ces exigences : il reprend 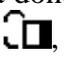, en rapprochant davantage les traits courbes qui représentent index et pouce, et le combine avec le glyphe 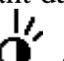, qui décrit la position des trois autres doigts. Nous obtenons

donc un glyphe comme 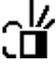 (mais qui n'existe pas dans l'IMWA) beaucoup plus proche que ce que l'auteur voulait représenter. Avec le glyphe 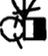 reste la différence de la position des doigts moyen, annulaire et auriculaire, qui n'est pas cohérent avec les autres exemples présents dans l'IMWA, mais qui ne suscite pas de problèmes de compréhension.

Nous retrouvons, en plus complexe encore, la même situation dans le glyphe  de l'exemple (d). L'auteur voulait représenter un hochement semi-circulaire de la tête, ce qui n'est pas possible avec l'IMWA2004, puisque dans ce système seulement les hochements rectilignes, comme , sont admis. Pour parvenir à son but, le scripteur à donc pris le glyphe , normalement utilisé pour représenter un mouvement semi-circulaire de la main, et l'a adapté en le réduisant et en le plaçant au dessus du rond représentant le visage, c'est-à-dire le lieu où sont normalement placés les mouvements d'oscillation de la tête.

Un élément particulier de ce nouveau glyphe est que la nécessité de sa création n'a pas été sentie exclusivement par les membres du SLDS ; en effet, si nous allons chercher dans l'ISWA2008, c'est-à-dire la version consécutive à l'IMWA2004 utilisée par nos auteurs, nous trouvons le glyphe conventionnel , qui est exactement le glyphe que l'auteur avait voulu créer (rappelons qu'aucun membre du SLDS ne connaissait, à l'époque, les versions suivantes de SW).

Nous sommes donc face à un processus qui permet à des auteurs avec une grande compétence en SW de repérer les éléments des différents glyphes qui véhiculent telle ou telle information, et de les combiner pour obtenir de nouveaux glyphes qui puissent satisfaire leurs exigences de représentation.

Il est aussi possible de trouver dans les textes écrits par nos auteurs des glyphes qui ne ressemblent en rien aux glyphes existants. C'est le cas du glyphe non conventionnel  présent dans l'exemple (d), dont l'explication nécessite de faire préliminairement le point sur le modèle sémiologique (Cuxac, 2000), cadre de référence théorique dans lequel s'inscrivent les recherches des membres du SLDS.

Dans le discours en LS il est possible de trouver deux types d'expressions référentielles, reconnues par tous les chercheurs mais différemment classifiées et nommées, selon le cadre théorique de référence (pour une discussion critique, voir Cuxac, 2000 ; Antinoro Pizzuto *et al.*, 2008 ; Antinoro Pizzuto *et al.*, in press ; Cuxac & Antinoro Pizzuto, 2010 ; Garcia 2010 ; Garcia & Derycke, 2010a) ; ces deux unités, en accord avec le Modèle Sémiologique et la proposition terminologique de Cuxac & Antinoro Pizzuto (2010), sont nommées *Unité Lexématique* (UL) et *Structure de Grande Iconicité* (SGI).

Les UL sont souvent définies comme *signes frozen* ou *standards*, sont assimilables aux mots des LV et sont souvent incluses dans les dictionnaires de LS ; pour de nombreux chercheurs, ces unités sont les seules à avoir une valeur linguistique. Les SGI sont, par contre, des unités référentielles complexes, dotées de traits fortement iconiques et multilinéaires ; elles sont très présentes dans les discours – surtout narratifs – en LS mais sont exclues des dictionnaires, car elles sont considérées comme des signes fortement idiosyncratiques et qui ne peuvent pas être lemmatisés (voir, par exemple, Johnston, 2008). Une des différences les plus visibles entre ces deux types d'unités est le regard : il est marqué (et donc dirigé vers une portion de l'espace de signation, qui est activée par le regard même) pour les SGI, alors qu'il est non marqué pour les UL (et donc dirigé vers l'interlocuteur du signeur).

Une autre précision importante est que, en SW, seuls les éléments qui semblent essentiels au locuteur pour la compréhension du signe sont reportés dans sa représentation ; ainsi, si la position des épaules est considérée comme pertinente - dans la Fig. 2, voir l'exemple (c) où les épaules sont légèrement inclinées à gauche – elle sera indiquée mais, le cas échéant, sa présence est vue comme une surcharge inutile d'information.

Revenons maintenant à l'exemple (d), qui montre la capacité des utilisateurs de SW à modifier le système non seulement pour des exigences de précision dans la représentation des formes, mais aussi pour annoter des phénomènes linguistiques. Le glyphe ⌒, que nous retrouvons la première fois dans Di Renzo (2006), représente le regard et indique que celui-ci est dirigé vers l'interlocuteur (d'où le « i »). Cette position étant considérée par défaut, le regard n'est normalement pas indiqué dans ce cas ; or nous sommes face à un glyphe qui *marque* l'existence d'un regard *non marqué*, c'est-à-dire la présence d'une UL et non d'une SGI. Ce choix, que nous ne retrouvons pratiquement qu'en transcription, indique, de la part de l'auteur, une conscience du rôle linguistique du signe qu'il est en train de représenter, conscience qu'il explicite par l'utilisation de ⌒, levant ainsi tout doute sur l'interprétation du signe en question.

Nous sommes toutefois face à un glyphe qui ne semble avoir rien en commun avec les autres glyphes du regard ou des yeux, comme ⌒ ou ⌒. Bien que moins compréhensible que le passage de ⊥ à ⊥ nous sommes tout de même face à une réélaboration sur la base des glyphes existants, puisque l'emplacement du glyphe ⌒ est situé en haut du cercle représentant le visage, comme les autres glyphes des yeux. Nous sommes donc en présence d'un glyphe non conventionnel dont le sens est, certes, moins transparent que pour les autres cas, mais qui reste toutefois cohérent avec le reste du système.

La flexibilité permise par l'écriture manuelle peut comporter aussi un changement des glyphes sur la base non d'une exigence de représentation, mais d'une facilité de réalisation : c'est le cas présent, toujours dans la configuration ⌘, dans l'exemple (b), qui représente les mains vues de côté et sur le plan horizontal.

En premier lieu, nous remarquons que le point d'attache entre les doigts est traversé par une barre ⌘. Or selon les règles de SW, les mains sur le plan horizontal sont représentées avec une séparation entre la paume (ici représentée par un carré blanc et noir) et les doigts (3 traits droits et 2 arrondis). En suivant cette règle, ces glyphes devraient être dessinés ⌘ et non ⌘. Nous pouvons donc formuler une hypothèse sur les différences entre les processus manuels, mais aussi mentaux, qui portent à la composition de ⌘ et de ⌘ (Fig. 3).

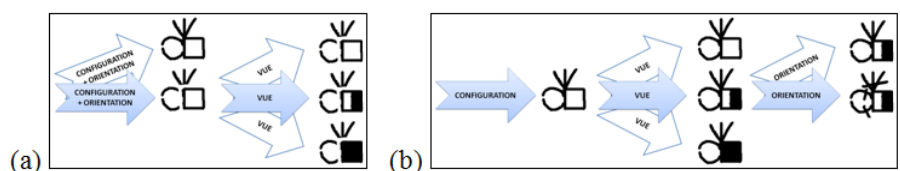

*Fig. 3 – Hypothèse sur les processus qui portent à la représentation des glyphes ⌘ (a) et ⌘ (b).*

Dans le cas de ⌘, le scripteur doit opérer dès le début, et en même temps, le choix de la configuration ( ⌘ ) et de l'orientation (plan vertical ⌘ ou horizontal ⌘ ) pour ensuite décider de la vue (paume ⌘, côté ⌘ et dos ⌘ ) ; dans le cas de ⌘, les trois choix sont opérés séparément, il y a donc ⌘ qui est marqué rapidement et par défaut, et ses autres caractéristiques qui sont ajoutées par-dessus ; ceci permettant, en outre, de corriger de façon plus aisée si l'on confond les deux plans. Mis à part le processus mental qui mène à la réalisation du glyphe, il faut aussi considérer que la solution non conventionnelle est plus simple à dessiner (il ne s'agit que de faire des traits et non d'interrompre des traits existants), et qu'elle est dotée d'une plus grande capacité à désambigüiser les deux plans, puisque les

traits 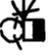 sont plus évidents que les « trous » de 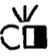, surtout dans le cas d'un scripteur n'ayant pas une bonne « calligraphie ».

Nous pouvons extraire, à partir des exemples que nous avons cités, les règles qui semblent régir la création spontanée de glyphes non conventionnels ou, plutôt, les règles qui font en sorte que ces glyphes soient acceptés par les autres utilisateurs de SW et ne soient pas sanctionnés comme des erreurs (sur le concept de faute d'orthographe en SW, cf. Bianchini 2014).

En premier lieu, ces glyphes doivent être cohérents avec le reste du système de glyphes de SW. Pour cela, il est préférable qu'ils reprennent des éléments déjà présents dans des glyphes officiels, pour qu'ils puissent satisfaire les règles gérant la représentation graphique de SW.

Ainsi, puisque tous les glyphes de torsion de l'avant-bras, comme 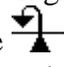, sont traversés par un trait horizontal (indiquant l'avant-bras), tout glyphe non conventionnel devra avoir recours à cette règle s'il veut être reconnu comme tel. De même, tout glyphe représentant le mouvement de la tête est indiqué par-dessus le cercle qui représente le visage, et seuls les nouveaux glyphes placés à cet endroit pourront être considérés comme des moyens non conventionnels de représenter ce phénomène.

Il s'agit ensuite de glyphes qui ont une utilité, c'est-à-dire qu'ils vont servir à représenter plus fidèlement un élément du signe, si cela n'est pas possible à travers des glyphes conventionnels. L'augmentation de précision peut être limitée, comme dans le cas de 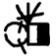 où 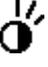 aurait probablement fait l'affaire, ou plus substantielle, comme dans le cas de 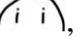, qui note un phénomène (le regard sur l'interlocuteur) qui n'est pas prévu par le système conventionnel.

En tous les cas, tout glyphe inventé doit respecter les règles de représentation graphique établies pour SW pour des glyphes analogues car c'est le seul moyen pour que ces glyphes non conventionnels soient facilement lisibles de la part de tout utilisateur expert de SW. La lisibilité est en effet l'un des critères dominants dans l'utilisation de SW : tout glyphe, mais aussi tout signe entier, illisible est sanctionné comme une erreur.

Les glyphes non conventionnels, s'ils respectent les qualités que nous venons d'énoncer (cohérence, utilité, lisibilité), peuvent aspirer à rentrer dans une version suivante de SW (ce qui a été le cas pour 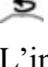 absent dans l'IMWA2004 mais qui est apparu dans l'ISWA2008 sous la forme 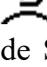). L'insertion de SW dans Unicode10.0 pourrait mitiger ce phénomène d'amélioration de SW, puisque, par principe, le Consortium ne reconnaît que des systèmes stables : toutefois, l'espace alloué à SW dans Unicode10.0 prévoit des cases vides qu'il serait possible de remplir avec de nouveaux glyphes, si reconnus d'une grande utilité. En tout cas, la présence de SW dans Unicode n'a de répercussions que sur sa version électronique, puisque Unicode n'influence pas l'écriture manuscrite.

## 3. Le SignWriting non conventionnel et l'écriture numérique : le SignMaker

Nous venons de voir que la liberté donnée par l'écriture manuelle n'est limitée que par les caractéristiques d'acceptabilité des glyphes non conventionnels. Toutefois, il est également possible d'écrire SW en format numérique et les instruments informatiques à disposition de l'utilisateur apportent de nouvelles limites que nous expliquerons dans cette section.

Actuellement, le logiciel officiel de SW – réalisé directement par Sutton et ses collaborateurs – est le *SignMaker*, dont nous avons déjà parlé rapidement dans les paragraphes précédents ; il s'agit d'une application web qui offre au scripteur la possibilité de représenter des signes en SW qu'il a normalement pré-composés sur papier (ce support étant beaucoup plus pratique que le logiciel).

Dans une session normale d'utilisation, le scripteur est appelé à chercher, parmi les glyphes qui composent l'IMWA ou - plus récemment - l'ISWA, ceux qui l'intéressent pour la composition du signes ; une fois qu'il les a trouvés, il doit les déplacer dans le *display du signe* (Borgia, 2010), équivalent numérique de la feuille blanche où sont agencés tous les glyphes nécessaires pour composer un signe. Ces glyphes peuvent par la suite subir des modifications, comme des changements d'orientation ou d'emplacement, pour aboutir à une représentation du signe qui soit la plus proche possible du signe que l'on voulait noter. Une fois l'assemblage terminé, il est possible de sauvegarder le travail effectué sous forme d'image ou de fichier *XML* contenant les noms et les coordonnées de tous les glyphes composant le signe.

Ce cadre nous permet de comprendre que l'utilisateur, obligé de se limiter aux glyphes contenus dans la version officielle de SW installée dans son logiciel, a une liberté inférieure à celle donnée par l'écriture manuelle où il peut avoir recours aux glyphes non conventionnels.

Pour avoir la même précision dans la représentation des glyphes que celle permise par l'écriture manuelle, il faut donc trouver un escamotage pour « forcer » le SignMaker. Dans ce logiciel, en effet, l'éventualité que l'utilisateur veuille utiliser des glyphes non conventionnels ne semble pas avoir été prise en compte. Les membres du SLDS ont donc recours, pour cela, à deux stratégies : l'évitement du non conventionnel, ou la composition à partir des glyphes officiels (Bianchini, 2012).

Dans le premier cas, nous sommes confrontés au sacrifice de la précision de représentation en faveur d'une plus grande simplicité de composition; nous trouvons donc, par exemple, la décision de remplacer 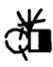 (Fig. 2b) par 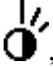, bien que, comme nous l'avons vu, les deux glyphes ne soient pas équivalents mais seulement ressemblants. Cette stratégie peut aussi comporter une surcharge graphique des glyphes, comme dans le cas de 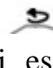 (Fig. 2d), que nous retrouvons représenté par 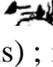 (par un membre du groupe qui essaye d'utiliser exclusivement des glyphes officiels) ; il est toutefois important de noter que cette application peut elle-même être considérée comme une utilisation non conventionnelle, puisque l'union d'un si grand nombre de glyphes dans cet espace est impossible avec le logiciel SignMaker (l'union des glyphes conventionnels 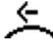 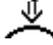 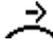 ne donnerait pas 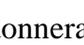 mais 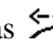, tout à fait illisible). La stratégie d'évitement est donc difficile à appliquer et limite les possibilités expressives de SW.

Dans le deuxième cas, le scripteur utilise toujours des glyphes officiels de SW, mais en regardant exclusivement leur forme et non leur fonction au sein de SW.

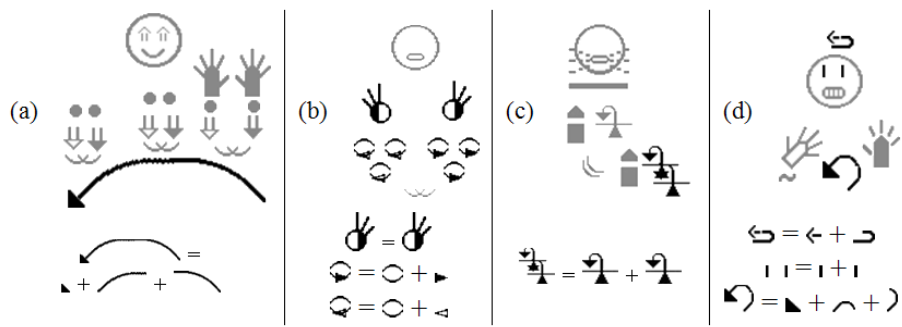

*Fig. 4 – Représentation numérisée de signes en SignWriting.*
*Les glyphes non conventionnels (en noir) sont le fruit de la composition de glyphes officiels ou dérivent d'une stratégie d'évitement du non conventionnel.*

Les exemples de la Fig. 4, qui sont la numérisation la plus fidèle possible de ceux présents dans la Fig. 2, montrent un des problèmes liés à cette procédure. Le fait de devoir choisir deux ou trois glyphes pour pouvoir en composer un seul, comporte une forte dépense de temps, surtout si l'on considère que les glyphes sont classés sur la base de leur fonction et non de leur forme (critère de choix utilisé ici) et que la recherche de chaque glyphe comporte jusqu'à 3 sélections dans le logiciel. Même dans le cas de l'évitement, comme dans la configuration ☝ de l'exemple (b), il est nécessaire de regarder plusieurs glyphes pour choisir celui qui répond le plus aux exigences du scripteur.

Le deuxième problème, moins évident, est le forçage de la procédure de codification des glyphes. Nous avons dit que chaque signe est sauvegardé sous forme d'image ou de fichier contenant le nom et les coordonnées de chaque glyphe qui le compose, ce fichier permettant d'exécuter des procédures de recherche, par exemple, tous les glyphes de mouvement de la tête. Or, dans les résultats, le signe de l'exemple (d) ne comparaîtra pas puisque ⇋ est composé de glyphes qui n'ont rien à voir avec les mouvements de la tête ; le problème est plus grave encore si l'on recherche tous les mouvements des doigts, puisque ⇋ est composé de ← plus ⌐ et que ← est un glyphe dont la fonction est justement celle de représenter ce mouvement, donc (d) ressortira parmi les résultats.

## 4. La résolution des problèmes de SignMaker : SWift

Face à ces problèmes de SignMaker (mais aussi à de nombreux autres, surtout liés à la difficulté de recherche des glyphes – pour une étude critique voir Borgia, 2010 et Bianchini, 2012), nous avons entrepris la création d'un nouveau logiciel, plus ergonomique et laissant plus de liberté à ses utilisateurs : le *SWift – SignWriting Improved Fast Transcriber* (Borgia, 2010 ; Bianchini *et al.*, 2012a-b). Comme son prédécesseur, cette application web permet de rechercher et d'agencer des glyphes dans une feuille blanche virtuelle, afin de composer des signes en utilisant SignWriting ; toutefois, toutes les procédures de choix et composition des glyphes sont *deaf centered* (Pizzuto *et al.*, 2010), c'est-à-dire conçues pour satisfaire en premier lieu les exigences des utilisateurs sourds, en particulier à travers un fort recours au visuel et une faible utilisation de la langue vocale écrite, ici l'italien.

Pour le recours aux glyphes non conventionnels, le SWift fait appel à deux procédures : la prévention et la gestion effective du problème.

Dans le premier cas, celui de la prévention, nous avons augmenté le glyphe de l'ISWA, en ajoutant les glyphes présents dans l'IWSA2008, ce qui fait passer le nombre de glyphes de 29 276 à 35 023, en incluant par exemple ⇋ dont nos utilisateurs avaient montré la nécessité. S'ajoute à cela un autre système de prévention qui est le fruit d'une analyse minutieuse de l'ISWA2008, qui a permis de mettre en évidence les lacunes du système, et de les combler[7].

Par exemple, pour les mouvements, nous sommes partis du présupposé que tout mouvement exécutable sur le plan horizontal l'est aussi sur les plans vertical et sagittal, et vice-versa, règle qui n'est pas toujours respectée en SW (y compris dans ISWA2010). Ainsi, s'il est possible de réaliser le mouvement curviligne sur le plan sagittal allant vers l'avant ↱, et sa répétition deux ↱ et trois ↱ fois, nous considérons que ceci peut aussi avoir lieu sur le plan sagittal vers le haut ↰, et sur les plans horizontal ⌒ et vertical ⌒ ; or pour ces trois plans, seulement la première et le deuxième répétition sont acceptées par l'ISWA2008 officiel. Nous avons donc rajouté les glyphes ⟨img⟩, ⟨img⟩ et ⟨img⟩. La Fig. 5 montre tous les cas où cela a été nécessaire.

---

[7] Un grand nombre des modifications que nous avons proposées se trouvent aussi dans la version ISWA2010 que Sutton a réalisé pour l'insertion dans Unicode. Nos propositions et celles de Sutton ont été élaborées de façon indépendante mais ont une large correspondance, ce qui montre comment les règles de créations des nouveaux glyphes soient « universelles » et non limitées au seul SLDS.

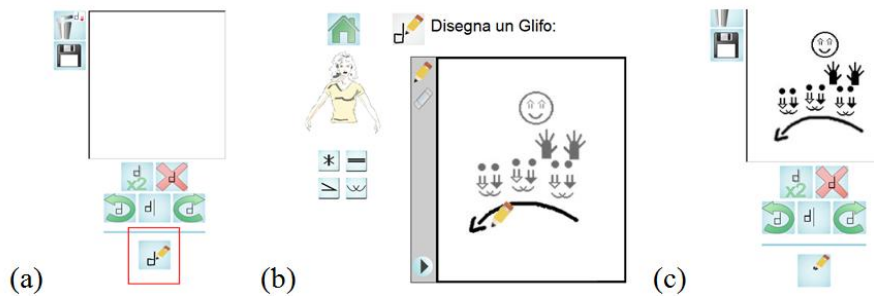

*Fig. 5 – Les incohérences dans la représentation du mouvement dans Sutton :*
*en blanc les glyphes existants, en couleur ceux que nous avons ajoutés car présents sur un autre plan*

Ce changement est le plus explicatif, mais non le seul que nous avons apporté à l'ISWA ; nous obtenons donc un nouveau ISWA comptant 47 930 glyphes (Bianchini, 2012 ; Bianchini et Borgia, 2012), ce qui permet aux signeurs d'utiliser des glyphes « conventionnellement non conventionnels ».

Dans le cas où l'utilisateur ne trouve pas le glyphe qu'il cherche dans l'ensemble étendu de glyphes, il peut recourir à une fonction du logiciel expressément conçue pour ces situations : l'aire de dessin à main levée (Fig. 6)

*Fig. 6 – Détails du fonctionnement de SWift : la sélection de l'icône du dessin à main levée (a)*
*ouvre une aire permettant de dessiner un glyphe non conventionnel (b) ;*
*une fois sauvegardé, il sera intégré au display du signe où sont agencés les autres glyphes*
*composant le signe que l'on veut représenter (c).*

Une fois que cette fonction est activée, s'ouvre une fenêtre pour le dessin à main levée, où se trouvent les glyphes déjà insérés ; c'est là que l'utilisateur, grâce à une tablette graphique ou une simple souris, peut dessiner le glyphe non conventionnel dont il a besoin. Cette fenêtre contient une version plus ample du *display* du signe, ce qui présente un avantage ergonomique certain ; la présence des glyphes déjà insérés permet par contre de respecter les proportions entre ceux-ci et le glyphe que l'on veut insérer.

Les instruments à disposition pour cette opération sont intuitifs et peu nombreux : un crayon et une gomme pour, éventuellement, corriger une erreur. Le recours à une tablette graphique, ou à un stylo numérique sur un écran *touch*, permet à l'utilisateur de dessiner avec un instrument proche d'un vrai stylo, mais une simple souris est suffisante pour obtenir un dessin précis.

Une fois que le glyphe non conventionnel a été inséré, il est sauvegardé dans la base de données du logiciel, procédure indispensable pour que celui-ci soit visible. Toutefois, nous ne voudrions pas que, vu la simplicité d'exécution et la liberté donnée par les dessins à main levée, les utilisateurs composent entièrement les signes à l'aide de la tablette graphique. Ce qui serait un avantage pour l'utilisateur serait en effet très problématique pour les analyses statistiques des représentations numériques des glyphes, fonctions indispensables à l'instrument de complétion automatique des signes qui est intégré au SWift et dont nous ne pouvons, pour des raisons d'espace, parler dans cet article (cette fonction est visible dans Borgia,°2010).

Pour éviter cette dérive, nous avons décidé de limiter l'accès aux glyphes non conventionnels aux utilisateurs : ils peuvent les insérer, les voir dans des signes déjà composés, mais ne peuvent pas les réutiliser pour créer un nouveau signe. Les glyphes insérés manuellement seront par contre à disposition des développeurs du logiciel et des chercheurs, pour des futures améliorations de SWift et pour l'élargissement de l'ensemble des glyphes possibles.

L'aire de dessin à main levée est aussi la base pour une nouvelle fonction de SWift, qui a été développée par Borgia dans sa thèse (Borgia, 2015), et qui devrait permettre de simplifier le processus d'insertion des glyphes : la recherche des glyphes conventionnels à partir de dessins à main levée, comme cela se fait déjà pour les écritures alphabétiques des langues vocales, à travers les systèmes de reconnaissance optique des caractères (O.C.R.). Cette fonction aura aussi l'avantage de suggérer aux utilisateurs la présence de glyphes conventionnels qu'ils ne connaissent pas, réduisant ainsi le recours aux glyphes non conventionnels mais redondants par rapports aux premiers. L'algorithme nécessaire à cette fonction a bien été développé (et est présenté dans le détail dans la thèse en Informatique de Borgia [2015]) mais il n'a malencontreusement pas encore été inséré dans SWift.

## 5. Conclusions

Dans cet article nous avons analysé un phénomène spécifique de l'utilisation de SignWriting : la création, de la part de sourds utilisateurs experts de SW, de nouveaux glyphes qui permettent une plus grande précision dans la représentation des signes. Ce phénomène permet la mise à jour continuelle de SW, qui se base sur les requêtes d'ajustement d'utilisateurs comme ceux avec lesquels nous avons travaillé.

De plus, nous avons aussi mis en évidence l'utilisation de cette implémentation de SW pour trouver des moyens d'annoter des phénomènes linguistiques, comme le regard rivé sur l'interlocuteur, qui indique l'utilisation d'une Unité Lexématique et non d'une Structure de Grande Iconicité.

Face à cette capacité des utilisateurs de SW de plier ce dernier à leurs exigences de représentation, les technologies numériques ne semblent pas suivre le rythme, limitant fortement la tendance naturelle des utilisateurs sourds à ajuster le système pour le rendre plus performant. Il est donc nécessaire de continuer sur la voie inaugurée par le SWift, qui se base sur SW en prenant en compte les exigences réelles de ses utilisateurs et non pas uniquement les caractéristiques techniques du système.

*Bibliographie*

*Sitographie*

La dernière consultation de tous les sites a été effectuée en janvier 2018.

[1] *History of SignWriting*. http://www.signwriting.org/library/history/hist003.html
[2] *SignMaker (version 1.6)*. http://www.signbank.org/SignPuddle1.6/signmaker.php
[3] *ISWA 2008*. http://www.movementwriting.org/symbolbank/ISWA2008/
[4] *ISWA 2010*. http://www.signwriting.org/lessons/iswa/
[5] *ISWA 2004*. http://www.movementwriting.org/symbolbank/IMWA2004/
[6] *Unicode10.0.* http://www.unicode.org/versions/Unicode10.0.0/Colophon.pdf